# Research on Personal Credit Risk Assessment Methods Based on Causal Inference

Jiaxin Wang[1]   YiLong Ma[2]

Causal inference originated from research in disciplines such as economics and biostatistics. With the development of artificial intelligence, causal inference is experiencing a resurgence of vitality, accompanied by new challenges. One particularly critical issue is the continuing weakness in its mathematical foundation, reflecting significant disputes among humans regarding the definition of causal relationships. To address this problem, a seminar titled "Foundations and New Horizons for Causal Inference," hosted by mathematician Oborwofahe, was convened in 2019 [1], aiming to unify existing methods and mathematical foundations, and facilitate the exchange of ideas across different domains. The seminar brought together top researchers from various fields, including artificial intelligence, biostatistics, computer science, economics, epidemiology, machine learning, mathematics, and statistics. In the same year, a comprehensive review of causal research was jointly authored by K. Kuang and others, covering various aspects of causal inference [2]. This paper includes K. Kuang's "Estimating Average Treatment Effects: A Brief Review and Others," Lian Li's "Attribution Issues in Counterfactual Reasoning," Lian Li's "Yule-Simpson Paradox and Alternative Paradox," Zhi Geng and Lei Xu's "Causal Potential Theory," Kun Zhang's "Discovering Causal Information from Observational Data," Kun Zhang's "Formal Argumentation in Causal Inference and Explanation," Beishui Liao and Huaxin Huang, Peng Ding's "Causal Inference of Complex Experiments," Wang Miao's "Instrumental Variables and Negative Controls in Observational Studies," and Zhichao Jiang's "Causal Inference of Interference." Yao, L., and others [3] reviewed existing methods for estimating causal effects and related assumptions. Their article mentioned matching methods, the Re-weighting method, and stratification methods for estimating causal effects, and discussed three causal assumptions: the unconfoundedness assumption, the positivity assumption, and the consistency assumption, among which the positivity assumption lacks a clear definition in big data. Tyler J. VanderWeele and others [4] questioned the boundary between the definition of causal effects and the definition of causal relationships. The article mentioned that the potential outcome framework only provides a set of sufficient conditions for defining causal estimation values, but some important issues are not suitable for the potential outcome framework. Currently, the definition of causal relationships remains undecided.

Currently, research on causal relationships can be broadly categorized into three directions: the definition of causal relationships, methods for estimating causal effects, and the application of causal inference + "X." The progress of any of these three research directions can contribute to the deconstruction of causal relationships. The relationship among these three directions is illustrated in the following figure:

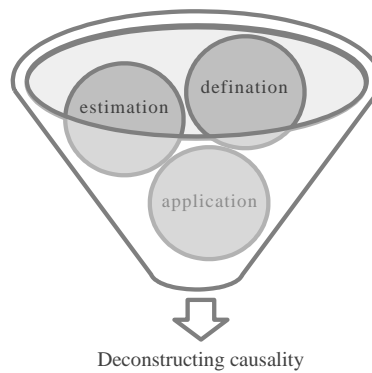

Deconstructing causality

1. Guilin University Of Electronic Technology
2. Northeastern University
Corresponding author: Yilong Ma 825133819@qq.com

# 1 Definition of causation

## 1.1 Philosophical definition

Since the scope of this paper is focused on the scientific methods of causal relationships, in this section, we will only discuss the definition of causal relationships in Western philosophy. The concept of causality as a fundamental notion in Western philosophy can be traced back to ancient Greece, but extensive discussions about it emerged in modern times. For example, the "Atomistic Theory" in ancient Greek philosophy emphasized that the interactions between atoms at the unobservable level are the causes of various phenomena at the observable level [5]. Aristotle's famous "Four Causes" theory, starting from the decomposition of subjects and predicates, established the notion of causality [6]. After Hume's reexamination of the problem of causality, many philosophers and mathematicians showed great interest in this issue once again.

Some terms used in the history of philosophy that need to be distinguished from "causal relationships" include "causal inference," "causality," and "causal laws." Among them, Hume often used "causal inference" as a term to introduce empiricism into discussions of causal relationships, focusing on exploring the cognitive evolutionary process when humans discover causal relationships [7]. "Causality" and "causal connection," as well as "causal verbs," are closely related, referring to the property exhibited by things that have causal relationships: causality is essentially equivalent to causal relationships, and causal verbs describe the verbs used to describe causal relationships between events, which are indicators in everyday language for humans to judge whether causal relationships exist [8]. In contrast to the former two, "causal laws" refer to the universal existence of causality in the universe, emphasizing generality and necessity [5]. The connotation of causal laws is broader than that of causal relationships, as causal relationships are a subset of causal laws.

The debate over causal relationships in modern philosophy revolves around the empirical versus rationalist debate, essentially focusing on the issue of the source of knowledge. The fundamental question is: "What ultimately provides an explanation for belief, reason, or experience?" or "Is the source of our ideas rational or empirical?" Rationalism holds that knowledge comes from innate ideas, and thus its characteristics can be summarized in two propositions: the mind possesses innate ideas, and beliefs about the world can be proven through reason. In contrast, empiricism denies the truth of these two propositions. Represented by Leibniz, the rationalists believe that every causal belief about the world can actually be generated through reason [10], and since we fully understand the concepts involved in beliefs, obtaining a concept means understanding the essence of the entity, that is, understanding all the attributes of the entity [11]. Locke, a pioneer of empiricism, believed that ideas come from experience, and the human mind has an "abstract" ability, by which humans derive abstract ideas from concrete ones [12]. Locke denied the existence of material entities while affirming the existence of mental entities, arguing that we cannot have knowledge about material existence but can only capture some sensory qualities of matter, thus unable to prove the existence of matter. There is a fundamental contradiction in Locke's philosophy: he assumed the existence of properties' substrata, the hypothetical material entities behind them, and believed that we can know the properties of objects, but matter itself does not exist. This contradiction was resolved in Hume's philosophy. Following the empiricist method in the history of philosophy and combining the "atomic" analysis method widely adopted in the development of modern natural science, Hume decomposed complex thoughts or concepts into the simplest and most primitive sensory experiences, namely impressions. These are the basic units of the smallest perceptions, and therefore they are the "clearest and most distinct." Hume's empiricist analytical method is generally referred to as "psychological atomism" [13].

The focus of Hume's philosophical research is epistemology, and the possibility and validity of knowledge are the main categories of his discussion. The problem of "causal relationships" is the core of Hume's epistemology and the "most important and distinctive part" of his philosophy. Causal relationships constitute Hume's theory of probable knowledge, and most of the knowledge of probability is constituted by causal reasoning. Since the proposal of causal relationships was to deal with Humean skepticism, it has become the central issue of Hume's philosophy. Hume's conception of causality appears as one of the three kinds of "complex ideas" (a classification method Hume used for ideas) and is the most important relationship in the discussion category. Therefore, we can infer that causality is also a kind of idea. This idea can be combined with "impressions" through the form of "imagination" to obtain "imaginary ideas," which serve as copies of impressions [14], so causality is also one of the three ways of combining ideas through imagination.

Based on the modern empirical thinkers' reflections, Hume proposed his unique conception of causality. Firstly, Hume put forward the theory of concepts, adhering to the principle of empiricism, believing that all perceptions come from the simplest sensory impressions [15]. However, Hume then encountered a problem: although causality can be explained as a product of simple impressions transformed by imagination and belief using the theory of concepts to interpret its structure and origin, it cannot explain what the essence of causality, namely necessity, is. We cannot find the source of necessity in impressions, so we cannot explain the essence of causality [16]. In Hume's philosophical thought, he found that the extreme of empiricism often leads to skepticism. However, for people's cognition of things and the source of knowledge, it cannot rely entirely on the viewpoint of skepticism because it cannot explain our common sense. To solve this problem, Hume proposed a new, unprecedented naturalistic attitude, attempting to explain the necessity of causality. He believed that the necessity of causality originates from human habits, and it is the emotions and psychological attitudes of human beings that prompt us to form beliefs that causes necessarily lead to results. Hume's irrationalistic epistemology opened up a completely new approach, which had a profound influence on modern philosophers.

In contemporary times, research on the problem of causality has become more meticulous and is significantly influenced by science, especially physics. Scientists attempt to express their views and opinions on this issue, while also emphasizing the importance of common sense in philosophical theory. The theories they propose provide new paths for us to better understand the relationship between mind and body and the relationship between mind and matter. In the new Humean controversy, quasi-realists attempt to uphold Hume's naturalistic stance while also accepting some views of skeptics. They reduce causality to the projection of the human mind into the objective external world, which can better explain Hume's "moderate skepticism" attitude and has important guiding significance for future research on causality [17].

1.2 Mathematical definition

Hume, in his philosophical thought, only established a temporal correlation between cause and effect. Kant, on the other hand, attempted to explore this issue through an a priori approach, shifting the concept of causality from "natural" to "rational" [18]. However, Kant's a priori solution did not truly address the problem of the inherent connection of causality but merely shifted the issue of causality to the realm of the a priori. Hume, through an empirical approach, studied the concept of causality, placing it within the realm of empirical science, leading to the study of the probabilistic nature of causal relationships. Kant's a priori method, on the other hand, highlighted its indispensable logical status through the exposition of universally existing causal assumptions. Due to Kant's limited understanding of the concept of causality within the a priori category, as Miller's research into causal relationships progressed, a clear trend of differentiation between scientific and philosophical studies of causality emerged.

Since Mill, the concept of causality as a basic assumption of inductive reasoning gradually became the object of logical research, later introduced into the field of study by scholars such as Russell, Reichenbach, and Carnap, shifting the study of causal relationships from absolute necessity to relative possibility. As possibility encompasses inherent logical contradictions as a characteristic of the objective world, this research orientation evolved into the method of "bold conjecture" in Popper's work. Meanwhile, another line of research focused on quantitative studies in fields such as statistical physics, statistics, econometrics, biostatistics, and psychometrics. In the quantification of causal relationships, Galton's linear regression model represented a shift from strict physical models to statistical models; Pearson's correlation coefficient, Keynes's inductive correlation, and Granger causality based on predictive methods dominated the study of correlation in empirical science. However, when these research methods were applied to fundamental science, especially quantum theory, the study of causality faced challenges and difficulties in quantification. If qualitative research into the concept of causality fell into the dilemma between experience and a priori, quantitative research into causal relationships was constrained by the confusion between causality and correlation.

Modern statistics is formally established on the basis of correlation criteria. Turing Award winner Judea Pearl believes that the purpose of statistical analysis is to evaluate the distribution parameters extracted from samples of this distribution. With the help of these parameters, people can infer the correlation between variables and estimate beliefs or probabilities of past and future times. It is assumed that as long as the experimental conditions remain unchanged, these tasks can be well managed through standard statistical analysis [19]. Causal inference, however, is different; its purpose is not only to infer beliefs or probabilities under static conditions but also to infer the dynamic changes of beliefs under changing conditions. This difference implies that the definition of causal relationships is not the same as the definition of correlation. Without a distribution function telling us how the distribution will change when external conditions change, because probability theory does not consider how the distribution of one feature should co-vary with changes in the distribution of another feature. Judea Pearl believes this implies the difference between correlation and causality and also suggests that behind every causal conclusion there must be some causal assumptions that cannot be verified in observational studies [20].

Pearl mathematically defined intervention actions and ingeniously introduced the Do-operator, combined with directed acyclic graphs to define causal relationships [2], [33]. Pearl views probability measures as degrees of belief. Let V be the set of variables ordered according to causal hypotheses, and let P(v) denote the probability distribution over the ordered set of variables. For each variable $X_i$ in V, Pearl refers to its independent parental variables as Markov parents or direct causes, denoted as $PA_i$ of $X_i$. In other words, if $Q_i(2\ PA)$ is the set of parental variables for X, then no proper subset of $PA_j$ can maintain the equality $P(x_j|pa_i) = P(x_j)$ according to Pearl. He argues that there is a lack of tools in probability measures to distinguish between intervention variables and observational variables. Therefore, he defines the intervention X = x as do(a). do(aj) means deleting all arrows from PA pointing to X in the causal graph and replacing the values of the function ti(pa,uj) with the constant X = xj. Pearl refers to the mapping from each C in X to P(y|do(a)) as the causal effect of X on Y. For each do(pay\X), Pearl defines the mapping from (;pay\X) to P(y|do()) as the direct causal effect, where pay\X is the intervention operation removing the Markov parental variables of X from Y.

According to the causal definitions mentioned above, Pearl identified three fundamental elements that constitute complex causal graphs: chains, forks, and colliders. Any complex directed acyclic causal graph can be composed of these three basic structures [23]. Subsequently, Pearl introduced the crucial criterion for identifying causal relationships, known as the backdoor criterion. He demonstrated that if a set of intermediate variables, denoted as B, satisfies the backdoor criterion, then the causal effect of A on Y can be estimated from observational data instead of intervention, i.e., P(y|do(a)) => P(y|z, b)P(z) [24]. Based on the backdoor criterion, Pearl further argued that B is the minimal set of variables that blocks all paths from A to Y. If B can block all backdoor paths from B to Y, and there are no backdoor paths in the paths from A to Y, then P(y|do(c))=P(y|a,b)P(a|c)P(c). This criterion is known as the front-door criterion [25].

1.3 Causality in social science problems

In the study of many significant issues in social science, the surgical precision of experiments, as advocated by Pearl, is often constrained by ethical or legal considerations. This problem is also a hot topic of debate in the definition of causal relationships. In 2021, the Nobel Prize in Economics was awarded to three economists, David Card, Joshua Angrist, and Guido Imbens, for their research on causal inference problems in natural experiments. Among them, labor economist David Card was awarded half of the Nobel Prize in Economics. He used natural experiments to analyze the effects of minimum wage, immigration, and education on the labor market [26]. His research since the early 1990s challenged traditional views and brought numerous new analyses and insights. The results showed that raising the minimum wage does not necessarily lead to a decrease in job opportunities. He made people aware that the income of those born in a country can benefit from new immigrants, while earlier immigrants are at risk of negative impacts. At the same time, he also made people realize that school resources are much more important for the future success of students in the labor market than previously imagined.

However, interpreting data from natural experiments is challenging. For example, extending compulsory education by one year for one group of students (as opposed to another group) will not necessarily affect each individual in that group in the same way. Some students will continue their education regardless, and for them, the value of education often cannot represent the entire group. So, is it possible to draw any conclusions about the effects of an extra year in school? In the mid-1990s, Joshua Angrist and Guido Imbens addressed this methodological issue and demonstrated how precise conclusions about causal relationships can be drawn from natural experiments. As a result, Joshua Angrist and Guido Imbens shared the other half of the Nobel Prize in Economics [27], [28].

2 Related concepts, methods and theoretical basis
2.1 Related concept
2.1.1 Causal inference

Causal inference is the process of determining the independent actual effects of specific phenomena that constitute part of a larger system. The primary distinction between causal inference and correlation prediction is that causal inference analyzes the response of the outcome variable when the causal variables change. (Pearl, Judea (1 January 2009). "Causal inference in statistics: An overview" (PDF). Statistics Surveys. 3: 96–146. doi:10.1214/09-SS057. Archived (PDF) from the original on 6 August 2010. Retrieved 24 September 2012.^ Morgan, Stephen; Winship, Chris (2007). Counterfactuals and Causal inference. Cambridge University Press. ISBN 978-0-521-67193-4.) People often view the task of causal inference as an inductive game between scientists and nature. Nature possesses stable causal mechanisms, which, at a detailed descriptive level, are deterministic relationships between variables. In natural experiments, some variables are unobservable. According to Pearl's causal theory, these mechanisms are organized in the form of directed acyclic graphs (DAGs) (J. Pearl, "Causal Diagrams for Empirical Research" Biometrika,82(4), 669--710, December 1995.).

Definition 1: A causal structure of a set of variables V is a directed acyclic graph (DAG) in which each node corresponds to a variable in V, and each link represents a direct causal functional relationship between the corresponding variables.

Causal structures serve as an effective way to describe how each variable is influenced by its parent variables. This definition implies the assumption that nature can freely impose any functional relationships between each outcome and its causes and then disrupt these relationships by introducing arbitrary disturbances.

Definition 2: A causal model $M=(D,P)$ consists of a causal structure $D$ and a parameter set P that is compatible with $D$. The parameter set P assigns a function $C_i=f_i(pA_i,U_i)$ to each variable $X_i$, and assigns a probability value $P(U_i)$ to each $U_i$, where $PA_i$ is the parent variables of $X_i$ in $D$, $U_i$ is a random disturbance according to the distribution $P(U_i)$, and all $U_i$ are mutually independent.

In this definition, Pearl provides a mathematical definition of causal graphs. Since the causal relationships are described by directed acyclic graphs, the definition implies two assumptions. Firstly, it assumes that scientists directly obtain distributions rather than obtaining events sampled from distributions. Secondly, it assumes that observed variables rather than some aggregations of them appear in the original causal model in a meaningful sense. This is because aggregations of variables or events would result in bidirectional paths between these aggregates, and directed acyclic graphs would be unable to describe such relationships with bidirectional cycles.

Influenced by the principle of Occam's razor, Pearl proposes a standard paradigm of scientific induction, where if we find a theory that is as consistent with the data as the existing theory but less complex, then we have reason to exclude the existing theory. The theory selected through this process is called the minimal theory.

Definition 3: If there exists a directed path from X to Y in every minimal structure consistent with the data, then variable X has a causal influence on variable Y.

Definition 4: A latent structure is defined as L = (D, O), where D is the causal structure on X and O is a set of observed variables.

Definition 5: A latent structure L = (D, O) is preferred over a latent structure L' = (D, O) (denoted as L ≻ L') if and only if D' can simulate D on O, that is, if and only if for every D, there exists a D' such that P_O(D, O_D) = P_O((D, D')). Two latent structures are equivalent, denoted as L ≡ L', if and only if L ≻ L' and L' ≻ L.

Definition 5 evaluates the preference for simplicity based on the expressive power of the structure rather than its syntactic description.

Definition 6: A latent structure L is minimally relative to a class of latent structures if and only if there is no structure in the class strictly preferred over L, that is, if and only if for every L' ∈ C, whenever L' is preferred, then L = L'.

Definition 7: A latent structure L = (D, O) has a distribution consistent with O if D can accommodate a model that generates P, that is, if there exists a parameterized D such that P_O((D, D')) = O.

Definition 8: Given P, variables C have a causal influence on variables E if and only if there is a directed path from C to E in every minimal latent structure consistent with P.

The above is currently the least controversial definition of causal inference, which is based on the semantic formulation of Occam's Razor principle.

2.1.2 Multi-angle data fusion

Data fusion, also known as information fusion, was initially applied in the military field. It is the study of using various effective methods to transform information from different sources and different time periods into information that can provide decision-makers with decision-making basis (Hua Bo-lin, Li Guang-jian. Theoretical and applied exploration of multi-source information fusion in the big data environment[J]. Library and Information Service, 2015, 59(16):5-10.). It refers to the process of combining data from different sources, forms, or types to obtain more comprehensive and accurate information. This process involves the integration, transformation, and analysis of multiple data sources to provide better support for decision-making, problem solving, or insight.

The objectives and explanations of data fusion are shown in the table below:

| Data fusion target | Target interpretation |
| --- | --- |
| Improve data integrity and reliability | Bringing together data from different sources can compensate for the limitations of each data source and improve data integrity and reliability. |
| Enhanced support for decision making | Combined analysis of multiple data sources can provide more comprehensive information to support a better decision making process. |
| Discover hidden associations and patterns | Combining information from different data sources can reveal hidden associations, trends, and patterns, providing deeper insights. |
| Reduce false positives and false negatives | By synthesizing information from multiple data sources, false positives and missed positives can be reduced, and the accuracy and reliability of decision making can be improved. |

This paper uses data fusion at the feature level, classifies data according to several dimension consideration criteria according to the business practice involved in the demonstration, and generates derived features from the classified features.

2.1.3 Natural experiment and controlled experiment

In causal inference, the definition of causal relationships is limited by whether the research environment is intervenable, so it is crucial to distinguish between natural experiments and controlled experiments in the inference process. When intervention is not feasible, in some cases, specific methods can bypass intervention and infer causal relationships solely through natural experiments, although in most cases bypassing intervention is not possible.

A natural experiment refers to an experiment in which the subjects are in an environment controlled by factors other than the researcher and cannot be controlled by the researcher (Friedman, G. D. Primer of Epidemiology 2nd. New York: McGraw-Hill. 1980. ISBN 978-0-07-022434-6). Natural experiments are observational studies rather than controlled studies.

A controlled experiment, also known as a controlled trial, refers to an experiment in which the researcher can intervene in the independent variables contained in the environment during the experiment to study their relationship with other non-independent variables. Pearl's definition of the Do operator for causal relationships is also developed through controlled experiments, as he believes that only surgical-like experiments can truly explore the mystery of causal relationships.

2.2.1 Random survival forest and its performance evaluation method

Survival analysis is an important field in statistics used to study the probability of an individual's survival within a specific time period, widely applied in various fields such as medicine, biology, and economics. In survival analysis, we are concerned with the time an individual experiences from a certain initial point to a survival endpoint, which could be survival time, unemployment duration, recovery time, and so on.

Survival analysis models can be classified into three main categories: semi-parametric models, parametric models, and non-parametric models. Semi-parametric models are a class of models between parametric and non-parametric models, with the most famous being the Cox proportional hazards model. The advantage of the Cox model lies in its flexible modeling of the baseline hazard while retaining strong mathematical form, making it widely used in practical applications. It allows us to examine differences in survival functions among different individuals by introducing covariates without making strong assumptions about the specific form of the baseline hazard function.

Parametric models, on the other hand, make specific assumptions about the form of the baseline hazard function, with common examples including the Weibull model and the exponential model. The Weibull model, by introducing a shape parameter, can flexibly adapt to different shapes of survival curves, thus having certain advantages in describing different types of survival data. The advantage of parametric models lies in their explicit assumptions about the shape of the survival curve, making the model more interpretable.

Non-parametric models, such as the Kaplan-Meier curve, do not make specific assumptions about the shape of the baseline hazard function. The Kaplan-Meier curve is an empirical distribution function, estimating the survival function by the cumulative proportion of surviving individuals over observed time. The advantage of this approach lies in its flexibility when facing complex and diverse survival data, as it does not have strong prior assumptions about the distribution shape.

Semi-parametric survival analysis models are a class of statistical models between parametric and non-parametric models. One of the most famous models is the Cox proportional hazards model, proposed by statistician David R. Cox in 1972. The Cox model is a semi-parametric model because it does not make specific assumptions about the shape of the baseline hazard function but allows the introduction of covariates to explore differences in survival functions among different individuals.

The fundamental assumption of the Cox proportional hazards model is that the risk of survival time is the product of the baseline hazard function and a function of covariates. Specifically, for individual i, their hazard can be represented as:

$$h_i(t) = h_0(t) \exp(\beta_1 x_{i1} + \beta_2 x_{i2} + \ldots + \beta_k x_{ik})$$

among, $h_i(t)$ is the risk of individual i at time t, $h_0(t)$ is the underlying risk function, $\beta_1, \beta_2, \ldots, \beta_k$ is a covariate $x_{i1}, x_{i2}, \ldots, x_{ik}$ Coefficient.

The advantage of the Cox model lies in its lack of specific assumptions about the shape of the baseline hazard function, making it more flexible in handling survival data. Additionally, it can accommodate right-skewed, left-skewed, and symmetric survival curves, making it suitable for various types of survival data. In practical applications, researchers often use the Cox model to study the impact of covariates on survival time. By estimating the regression coefficients of the model, we can understand the relative influence of each covariate on survival risk without making specific assumptions about the shape of the baseline hazard function.

Based on the current application trends in the industry, semi-parametric models are the most suitable class of survival analysis models for default prediction. With the maturity of ensemble learning methods, the performance of survival analysis models can be further improved using the principles of ensemble learning. Below are brief descriptions of two ensemble learning methods:

Bagging and Boosting are two main strategies of ensemble learning. Assuming we have a dataset consisting of n samples, each represented as (Xi,yi), where Xi is the input features and yi is the corresponding label.

The idea of Bagging is to, for each round t, generate a training subset Dt of size n through bootstrap sampling, where samples are drawn with replacement. Train a base model ht on each training subset Dt, and then combine all the predictions of ht by averaging or voting. For regression problems, it can be represented as:

$$\hat{y}(x) = \frac{1}{B} \sum_{b=1}^{B} M_b(x)$$

For classification problems, voting can be performed. Boosting begins by initializing the model and defining a base model *h0*. In each round *t*, the residuals of the model *ht* are calculated. A base model *ht* is trained using these residuals, and the predictions of each base model are combined according to certain weights. For regression problems, it can be represented as:

$$\hat{y}(x) = \sum_{t=1}^{T} \alpha_t M_t(x)$$

Overall, the fundamental idea of Bagging and Boosting is to build multiple base models and combine them to enhance overall performance. The main difference in their principles lies in the sampling method of the samples and the training strategy of the models. Boosting gradually corrects the model's predictions using the gradient descent approach, minimizing the weighted loss function to control generalization error. In contrast, Bagging relies on the Bootstrap sampling method to randomize the sample data and alleviate the overfitting issue caused by repeated sampling training. To explore the applicability of these two ensemble methods to survival analysis, this study designed comparative experiments to discuss which strategy is more suitable for improving the performance of survival analysis.

Due to the methodological foundation of statistics and machine learning theory utilized in this study, which constructs survival analysis models considering survival time, the task objective is to select covariates strongly correlated with the state variable. Therefore, it is necessary to not only test the accuracy of the model but also evaluate its calibration performance and whether the model can correctly determine the sequence of events. To comprehensively address these requirements, this study adopts three evaluation metrics: Harrell's concordance index (c-index), time-dependent AUC curve, and time-dependent Brier score.

Harrell's concordance index was first proposed by Professor Frank E. Harrell Jr. from Vanderbilt University in 1996. It calculates the agreement between the predicted order of events by the model and the actual occurrence order. Assuming we have a sample set where each sample contains an observation time and an event indicator (whether the event occurred), and the model assigns a predicted risk score to each sample. For any two samples i and j, if the model's risk scores can accurately reflect the order of their event occurrences, i.e., $t_i<t_j$ if and only if $\hat{R}(t_i)<\hat{R}(t_j)$, where $t_i$ and $t_j$ are the observed times of samples i and j respectively, and $\hat{R}(t_i)$ and $\hat{R}(t_j)$ are the model's predicted risk scores for samples i and j, then they are said to be concordant. The number of concordant pairs is denoted as $N_c$, and the number of discordant pairs where the model's risk scores are equal is denoted as $N_d$. Utilizing these statistics, Harrell's concordance index is calculated as $index=\frac{N_c}{N_c+N_d}$. The value of Harrell's concordance index ranges from 0.5 to 1, with values closer to 1 indicating better ability of the model to rank event occurrence times. When $index=0.5$, it means the model's ranking ability is equivalent to random guessing. C-index is a non-parametric method and is applicable in various survival analysis scenarios as it does not rely on specific assumptions about the distribution of survival times. It primarily focuses on the relative ranking of event occurrences by the model. Therefore, it is recommended to use other indicators such as survival curves to comprehensively evaluate the model's performance.

Receiver Operating Characteristic (ROC) curve is a tool used to evaluate the performance of binary classification models. The ROC curve plots the true positive rate (TPR) against the false positive rate (FPR) at different thresholds, where TPR is also known as sensitivity or recall, representing the model's ability to correctly predict positives. The formula for TPR is $TPR=\frac{TP}{P}$, where TP is the number of true positives and P is the total number of actual positives. FPR represents the rate of false positives and is calculated as $FPR=\frac{FP}{N}$, where FP is the number of false positives and N is the total number of actual negatives. In a binary classification problem, the model predicts based on probability or score outputs, and by setting a threshold, samples are classified as positives or negatives. The ROC curve changes the threshold and observes the changes in TPR and FPR. Each change in the threshold corresponds to a (TPR, FPR) point on the curve. In ideal conditions, the ROC curve will be close to the upper left corner, indicating high TPR and low FPR, meaning the model can effectively distinguish between positives and negatives. Here, we consider cumulative cases and dynamic situations at a given time point, which produces time-dependent AUC/time dynamic curves where the curve is plotted against time with the AUC value at that time point as the vertical axis.

The time-dependent Brier score is an extension of mean square error for right-censored data. Given a time point t, the Brier score is calculated as $BS_{Bc}(t) = \frac{1}{N}\sum_{i=1}^{N} BS_{ic}(t)$, where $BS_{ic}(t)$ is the predicted probability of remaining event-free for feature vectors prior to time t and is the inverse probability of being censored. The Brier score is commonly used for calibration assessment. If a model predicts a 10% risk of experiencing an event at a certain time, the observed frequency in the data should match this percentage for a well-calibrated model. Additionally, the Brier score is also a measure of discrimination: whether the model can predict risk scores that allow us to correctly determine the sequence of events.

2.2.2 Data fusion method

The methods for generating derived features through data fusion can be categorized into two types: constructing derived features based on statistical indicators and constructing derived features based on business meanings ([1] Tao Zhourong. Research on Portfolio Model of Personal Credit Risk Assessment [D]. East China Normal University, 2023. DOI: 10.27149/d.cnki.ghdsu.2022.003730.). The empirical research section of this paper relies on personal credit business in banks, so we adopt the method of linear regression to construct multiple original features into derived categorical features based on business meanings.

Given n samples with p features, , the corresponding linear regression model can be represented as follows: $y_i = \beta_0 + \beta_1 x_{i1} + \beta_2 x_{i2} + ... + \beta_p x_{ip} + \epsilon_i$

Where $y_i$ is the dependent variable, $x_{ij}$ represents the j-th original feature of the i-th sample, $\beta_0$ is the intercept, $\beta_j$ represents the multiplier for generating derived features from the p-dimensional original features, and $\epsilon_i$ is the fitting residual term.

2.2.3 Inductive Causation algorithm

When dealing with potential causal graph equivalent structures, this paper utilizes the built-in Inductive Causation (IC) algorithm in DoWhy. This algorithm takes a stable probability distribution generated by a potential directed acyclic graph as input and outputs patterns belonging to equivalence classes of the directed acyclic graph, i.e., subgraphs that are partially isomorphic to it.

Input: Stable distribution on variable set V  Output: Patterns compatible with D

1. For each pair of variables a and b in V, find the set S such that P(a, b|S) = P(a|S)P(b|S), i.e., a and b are independent in D given S. Construct an undirected graph G such that nodes a and b are connected if and only if set S does not exist.

2. For each pair of non-adjacent variables a and b with a common neighbor c, check if P(a, b|c) ≠ P(a|c)P(b|c): If true, continue; otherwise, add a directed edge pointing to c (i.e., a → c ← b).

3. In the obtained partially directed graph, attempt to direct as many undirected edges as possible based on the following two conditions: (a) any optional direction would create a new v-structure; (b) any optional direction would create a directed cycle.

Step 3 of the IC algorithm can be systematized in several ways, but to obtain maximal directed patterns from any pattern, the following four rules are required:

- Rule 1: If there exists an arrow a → c such that a and c are not adjacent, then direct b-c as a → c.
- Rule 2: If there exists a chain a-b-c, then direct a-b as a → b.
- Rule 3: If there exist two chains a-b-c and d-e-f such that c and d are not adjacent, then direct a-b as a → b.
- Rule 4: If there exist two chains a-b-c and d-e-f such that c and b are not adjacent and a

and d are adjacent, then direct a-b as a → b.

Repeated application of these four conditions is sufficient to direct all arrows shared by the equivalence classes of D. (Meek, 1995 C.Meek.Causal inference and causal explanation with background knowledge. In P. Besnard and S. Hanks, editors, Uncertainty in Artificial Intelligence 11, pages 403-410. Morgan Kaufmann, San Francisco, 1995.)

2.3 Correlation theory

2.3.1 Causal diagrams identify conditional theory

Theorem 1 (Do Operator Rules) Let G represent the directed acyclic graph corresponding to the defined causal model, and let P represent the probability distribution implied by this model. For any disjoint subsets of variables X, Y, Z, and W, Pearl proposed and proved the following rules (Pearl, 1995a J. Pearl. Causal diagrams for empirical research. Biometrika, 82(4):669-710, December 1995.): Rule 1 (Insertion/Deletion of Observational Variables): If X and Y are disjoint, then Rule 2 (Action/Observation Exchange): If X and Y are disjoint, then Rule 3 (Insertion/Deletion of Actions): If X and Y are disjoint, then where Z is the set of nodes in G that are ancestors of Z but not ancestors of W. Shpitser and Pearl proved the completeness of these three rules (Shpitser and Pearl, 2006a 1. Shpitser and J Pearl. Identification of conditional interventional distributions. In R. Dechter and T.S. Richardson, editors, Proceedings of the Twenty-Second Conference on Uncertainty in Artificial Intelligence, pages 437-444. AUAI Press, Corvallis, OR, 2006.). Theorem 2 (Galles and Pearl, 1998 D.Galles and J. Pearl. An axiomatic characterization of causal counterfactuals. Foundation of Science, 3(1):151-182, 1998.): Let X and Y be two variables in the semi-Markov model graph G. The identifiable sufficient conditions are satisfied if G meets one of the following four conditions:

1. There is no backdoor path from X to Y in G, i.e., .
2. There is no directed path from X to Y in G.
3. There exists a node set B that blocks all backdoor paths from X to Y, making Y identifiable. (In particular, when B consists entirely of non-descendants of X, it can be immediately reduced to .)
4. There exist two node sets and such that: (a) They block every directed path from X to Y (i.e., ). (b) They block all backdoor paths between and Y (i.e., ). (c) They block all backdoor paths between X and (i.e., ). (d) They do not activate any backdoor paths from X to Y (i.e., ). (If conditions (a)-(c) are satisfied and no node in is a descendant of X, then this condition holds.)

2.3.2 Category theory

The fundamental viewpoint of category theory is that using arrow diagrams can simplify many mathematical properties' proofs. Category theory arose in response to Bertrand Russell's set-theoretic paradox of self-reference, and the theory cleverly avoids the occurrence of Russell's paradox by utilizing unit morphisms.

Definition 9: A category C must satisfy four definitions and two properties:

1. A class $\mathrm{Ob}(\mathcal{C})$, whose elements are called objects of C.
2. For any $x, y \in \mathrm{Ob}(\mathcal{C})$, there exists a set $\mathcal{C}(x, y)$, whose elements are called morphisms from x to y.
3. For any $x \in \mathrm{Ob}(\mathcal{C})$, there exists an element $1_x \in \mathcal{C}(x, x)$, called the identity morphism of x.
4. For any x,y,z∈Ob(C), there exists a mapping: o:C(x,y)×C(y,z)→C(x,z)(f,g)↦gof;u called composition.

5. Unit law: For any x,y ∈ Ob(C), and for any f ∈ C(x,y) and Y ∈ C(z,y), we have f;1y=f=1z;f.
6. Associativity law: For any x,y,z,w ∈ Ob(C), and for any f ∈ C(x,y), g ∈ C(y,z), and h ∈ C(z,w), we have (gof)=(hog);of, and it can be written as hogof without ambiguity.

Objects that satisfy Definitions 1-4 and obey the unit law and associativity law are called categories.

Definition 10: Let there be four objects A, B, C, D in the category C, along with the arrows as shown in the diagram: A→fB→gC→hD. If f;g=g;f, then the diagram is said to commute.

Definition 11: Let C and D be two categories. A functor (or covariant functor) F:C→D must satisfy the following four conditions:
1. F:Ob(C)→Ob(D), meaning F maps each object in C to an object in D.
2. For any Y ∈ Ob(C), there exists a mapping F:C(X,Y)→D(F(X),F(Y)), which maps a morphism f:X→Y in C to a morphism F(f):F(X)→F(Y) in D.
3. For each X ∈ Ob(C), there exists 1F(X)=1F(X), meaning the identity morphism is mapped to the identity morphism.
4. For any X,Y,Z ∈ Ob(C), and any f ∈ C(X,Y) and g ∈ C(Y,Z), F(g∘f)=F(f)∘F(g). In other words, F maps commutative diagrams in C to commutative diagrams in D.

Definition 12: Let C and D be two categories. All functors from C to D and their natural transformations between them form a category called the functor category from C to D, denoted by Fun(C,D).

Definition 13: Let C and D be two categories, and F and G be functors from C to D. A natural transformation α from F to G, usually denoted as F⇒G, requires the following: For any morphism f:X→Y in C, there exists a commutative diagram:

$$\begin{array}{ccc} F(x) & \xrightarrow{F(f)} & F(y) \\ \downarrow \alpha_x & & \downarrow \alpha_y \\ G(x) & \xrightarrow{G(f)} & G(y) \end{array}$$

3 Theoretical proof of multi-angle data fusion causal inference based on category theory

3.1 Problem description

Any causal inference is conducted with a specific purpose in mind, which is reflected in the sequence of causal relationship discovery. We first observe at least two variables and then propose hypotheses about possible causal relationships between them. Therefore, before conducting causal analysis, we already have a set of hypotheses. Since there is no consensus on the sufficient conditions for causality in philosophy, the process of analyzing existing theories is actually about eliminating semantic interpretations that do not meet the necessary conditions for causality in various ways. In short, the entire process is not about affirming causal relationships but about negating non-causal relationships, and this process of causal inference is based on the exclusivity of causal relationship interpretations.

In fact, the direct definition of causal relationships is still a contentious issue. Although in experimental data, this paper adopts Pearl's interventionist definition. This definition is also the most widely used definition in current data science. However, intuitively, the author still feels that Pearl's interventionist definition is conceptually ambiguous. For example, when the human cultural background changes, cognitive representation data will change. When considering intervention operations in the experimental background of this paper, the causal graph structure, paths, and conditional probabilities will vary greatly. If interventions in the real world were possible, the results of interventions would also differ significantly. Therefore, this section boldly chooses category theory as a theoretical basis to explore another mathematical exposition of causal relationships.

Hume (1740) proposed five inferences about causal relationships: (1) Similar causes produce similar effects. (2) The difference in results between two similar objects must come from their differing characteristics. (3) When multiple different objects produce the same result, it must be through some common property we have discovered about them. (4) If an object increases or decreases with its cause, then that object should be considered a compound result. (5) If an object exists for a period without producing any results, then it is not the sole cause of that result.

To make the discussion clearer, this section further elaborates Hume's causal views using the language of category theory. In the language of category theory, we can use three key concepts to describe the aforementioned causal properties: objects, morphisms, and natural transformations, which correspond to causal representation, causal relationships, and transformations of causal relationships, respectively.

3.2 The causality view of Hume's philosophy

The most brilliant achievement of the British philosopher David Hume's life lies in his discussion of two questions, which the German philosopher Immanuel Kant referred to as the "Humean questions." These two questions can be summarized as the problem of induction and the problem of causality. Because of the inherent connection between these two issues, they are mutually reinforcing. Therefore, the discussion of Humean philosophical ideas by later scholars ultimately evolved into one question, namely the problem of causality.

Hume divided human perception into two concepts: impressions and ideas. The difference between them lies in the intensity and liveliness of the mind's stimulation and the entry into thought or consciousness. Hume believed that impressions are more lively and vivid than ideas. In the term "impression," Hume includes all sensations and emotions that first appear in the soul. As for ideas, Hume refers to the faint images given by sensations and emotions in thinking and reasoning, dividing them into simple ideas and complex ideas. There is a connection between ideas. Hume generalizes the nature of the transition of the human mind from one idea to another into three types: resemblance, contiguity in time and place, and cause and effect. In discussing the concept of causality, Hume encompasses two layers of meaning: one object being the cause of another object, and one object being the cause of the activity or movement of another object. After discussing the concept of causality, Hume continued to discuss the issue of the origin of causality. He asserted that the idea of causality must stem from some relationship between objects, which requires us to discover it.

3.2.1 Definition of causation

In Hume's theory of causality, the concept of causation is reducible. That is to say, we can consider causation as composed of some non-causal factors, and through these non-causal factors, causation can be reduced (J.A. Robinson, L.W. Kang. Two Definitions of Hume's "Causality" [J]. World Philosophy, 1997(2):61-65, 40.) Hume's definition of causation can be summarized in logical language as follows:

x is the cause of y if and only if the following three conditions hold:

1. There is a temporal succession between y and x.
2. The connection between x and y is constant.
3. x and y are contiguous in space.

Regarding temporal succession and spatial contiguity, Hume points out: "If an object exists in a complete state for a period of time without producing another object, it is not the sole cause of the other object, but must be supplemented by other principles that can propel it from an inactive state and enable it to exert its hidden potential. However, if the cause and its effect are simultaneous, then all causes and effects will be determined in this way, because any object among them will delay its action at an instant, thus preventing it from acting at the moment when it should have done so. Then it is not a proper cause. The result is no less than destroying the succession of causes we find in the world and completely eliminating time, because if a cause and its effect are simultaneous, and this result is simultaneous with its effect, and so on, then obviously there will be no succession phenomenon, and all objects must exist simultaneously" (Hume. A Treatise of Human Nature [M]. Translated by Shi Biqiu. Beijing: China Social Sciences Press, 2008, p.92.)

Regarding the constancy of connection, Hume emphasizes in "A Treatise of Human Nature": "Anything cannot act outside the time or place where it exists. Objects that are far apart sometimes seem to produce each other. Upon examination, it is often found that they are connected by a series of causes, which themselves are close, but when we cannot find such a connection in any particular case, we can still assume that such a connection exists" (Hume. A Treatise of Human Nature [M]. Translated by Shi Biqiu. Beijing: China Social Sciences Press, 2008, p.91.)

The author believes that during Hume's time (1711-1776), relativity had not yet emerged, and Hume was unaware of the contradiction between the concept of space-time and causality in physics, which is the limitation of Hume's causal view. The author, based on Henri Bergson's concept of duration (Henri Bergson. Creative Evolution [M]. Huaxia Publishing House, 1999, p.43.), proposes a new interpretation of Hume's concepts of time and space. Time should be qualitative rather than quantitative; it is an extension of continuous heterogeneous flow that is constantly surging forward. There is no difference in quantity and intensity between the various moments of duration; they permeate and merge with each other in nature, forming an indivisible and immeasurable whole. In it, the present always contains the past and carries it into the future; duration is the continuous progress of the past, gradually engulfing the future, and as it advances, it also expands. Space, on the other hand, is quantitative; it can be measured and perceived.

3.2.2 The source of causation

Hume believed that "causal relationships are discovered not through reason, but through experience." (Hume, A Treatise of Human Nature, translated by Guan Wenyun, Commercial Press, Beijing, 2008, p.28.) Hume regards experience as the basis for establishing causal relationships, where experience refers to the concept of causality formed through habit. Hume's view of causality is also based on the deconstruction of the process by which habits generate the concept of causality. Hume pointed out that "cause and effect are two perceptual objects, and causal inference transitions from the affirmation of one object to the affirmation of another. However, affirming one object while negating the other does not lead to self-contradiction, indicating that causal inference does not have the necessity usually attributed to it; causes do not entail their effects." Hume's hypothesis of the uniformity of nature, as pointed out here, is an irrational assumption. He believed that if causal inference were the result of rational thinking or domination, there would be a conscious reflection when inferring from causes to effects. In reality, however, the process of causal inference is unconscious. There is no pause between inferring from one object to another. Inferring from causes to effects often cannot be derived from a single impression but requires multiple repetitions of impressions. If causal inference were based on rational thinking, it would not need the constant conjunction of two types of objects, as constant conjunction does not add any new ideas beyond a single conjunction.

The author largely agrees with Hume's analysis of the origin of causality, considering causality as a result of habit. Hume mentioned, "The deconstruction of causal relationships can be beneficial to us by allowing our experiences to inform us, enabling us to transcend the senses and infer knowledge about the future. Otherwise, we would never know how to make our means achieve our ends, and would never use our natural abilities to produce any results." (Hume, A Treatise of Human Nature, translated by Guan Wenyun, Commercial Press, Beijing, 2008, p.28.) In the next section, the author will provide a mathematical abstraction of the above causal relationships and their origins.

3.3 Causal object definition based on category theory

3.3.1 Definition of causal objects

This section interprets the semantic interpretation of data as follows: "The value of variable XX of object A is XXX." Consistent with Hume's causal view, causal facts are defined as elements contained in the category class of causality.

The consideration of a class of causal questions by an individual or a group of individuals can be represented by a series of data. For example, the foundational logic of the AHP model, most representative and influential in expert experience methods, represents the experience of an expert or a group of experts in the form of a judgment matrix. The definition of causal objects requires consensual assumptions:

(1) The consideration of any causal question is not arbitrary, but based on a knowledge foundation, which is reflected in the relevant information encountered by the individual or group of individuals considering the question in reality.

(2) If a group rather than an individual attempts to address a causal question, there are certain consciousness connections among the individuals in this group.

(3) When a group or an individual within a group does not establish any consciousness connection with other groups or individuals outside of it, their views on a causal question will not change.

(4) In the process of solving a causal question for a group, the consciousness connections among groups and individuals within the group are transitive. Moreover, this transitivity is independent of the order when no new consciousness transmission occurs.

Definition 13: A class of causal questions is described as a category class $Ob(C)$, where each category represents an abstract form of data tables used to represent the thinking of an individual or a group about the problem. If there are connections among the consciousnesses of the individuals considering this class of causal questions, the category is called a causal category.

1. Based on assumption (1), for any two causal categories $x, y \in Ob(C)$, there exists a set $C(z,y)$ where its elements $f \in C(x,z)$ are called causal morphisms from one causal category to another. Causal morphisms represent the transformation of representation modes.

2. From assumption (3), for any causal category $x \in Ob(C)$, there exists an element $Idx \in C(x,x)$ called the causal identity morphism. The causal identity morphism can be intuitively understood as follows: in terms of time, when individuals or groups do not establish new connections with the outside world, there is no change in the representation of causal questions. The causal identity morphism can be understood as a process of self-representation.

3. From assumption (4), for any three causal categories x,y,z ∈ Ob(C), there exists a mapping o:C(x,y)×C(y,z)→C(x,z) where (f,g)↦g∘f, called the composite mapping, is referred to as the causal concept transmission mapping.

4. For any two causal categories x,y ∈ Ob(C), and any morphism f ∈ C(x,y), from assumptions (3) and (4), it follows that Idy∘f=f=f∘Idx. The causal category's unit law can be intuitively understood as follows: if there is a connection between the individuals or groups behind two representation modes, then this connection, whether it occurs before or after the causal identity morphism or self-representation process, does not affect the representation results.

5. From assumption (4), for any causal categories x,y,z,w ∈ Ob(C), and any causal morphisms f ∈ C(x,y), g ∈ C(y,z), h ∈ C(z,w), it holds that h∘(g∘f)=(h∘g)∘f, unambiguously denoted as h∘(g∘f)=(h∘g)∘f.

Definition 14: Let C be a causal category. x,y∈C are said to be isomorphic if and only if there exist two causal morphisms f∈C(z,y) and g∈C(y,z) such that g∘f=Idz. We denote g=f−1, called an inverse morphism of f, and f is referred to as a causal isomorphism between C and Y.

3.3.2 Definition of causation

Definition 15: Let C and D be objects in C, two causal categories. F is called a causal functor from C to D (implying a causal relationship between representations, which is a functional relationship) if and only if:

1. For any x in C, F(x) is defined in D, i.e., the functor maps objects to objects. Here, objects refer to specific representations within the abstract representation category.

2. For any z,y in C and f in C(z,y), F maps f to D(F(z),F(y)), i.e., the functor maps morphisms to morphisms.

3. For any x in F(Idx)=IdF(x), i.e., the functor maps identity morphisms to identity morphisms.

4. For any z,y,z in C, g in C(y,z), and f in C(z,y), there is $F(g \circ f) = F(g) \circ F(f)$.

Defining causality as a causal functor that belongs to two causal categories is inspired by Hume's philosophical view of causality. Hume claimed that causality exists only in the human psychological world. Intuitively, this can be understood as data being representations of psychological cognition, and for a given class of causal problems, these representations are abstracted from the psychological world of the individuals behind them. Therefore, causality is the relationship that allows this class of causal problems to exist and enables certain cognitions in the psychological world of individuals contemplating these problems to interact with each other.

3.3.3 Definition of causal transformation

Definition 16: Let C and D be two causal categories in Ob(C), and F and G be two causal functors (representing causality) from C to D. Then, a " is a causal natural transformation from F to G if and only if: "

1. For any X in C, there exists ηX in D such that ηX is a morphism from F(X) toG(X).

For any y and f in C, (y)F(f)=G(f)ηX.

This section employs the concept of causal natural transformations to describe how changes in the consciousness of a group of individuals when contemplating a causal problem, due to interaction with the external world, result in changes in the causal relationships within the problem.

Using the definition of natural transformations from category theory, this section defines such transformations as causal natural transformations. Causal natural transformations are also influenced by various factors such as the human social environment and advancements in human cognitive understanding. In summary, causal natural transformations provide an abstract definition of how causal relationships change.

3.4 Proof of data fusion theory based on category theory

3.4.1 Yoneda lemma and its intuitive understanding

The Yoneda Lemma is an important result in category theory, playing a crucial role in understanding the structure and properties of categories. It was proposed by the Japanese mathematician Nobuo Yoneda in 1954.

Yoneda's Lemma: Let C be a locally small category. For any $X \in C$ and $y \in C$, there exists a natural transformation from the functor $HomHomC(-,z)$ to the functor $HomC(-,y)$, given by $Nat(HomC(-,z), HomC(-,y))$, where $HomC(z,y)$ represents the set of morphisms from X to Y, and $Nat(HomC(-,z), HomC(-,y))$ represents the set of natural transformations from the functor $HomC(-,z)$ to the functor $HomC(-,y)$.

In other words, Yoneda's Lemma states: Let C be a locally small category, and let F be a functor from C to the category of sets, denoted as Set. For any $A \in C$, there exists a natural transformation $Nat(hA,F) = Hom(Hom(A,-),F)$, where $Hom(A,-)$ is the hom functor set, mapping A to a set of hom functors. This natural transformation is a bijection with $F(A)$.

Furthermore, when both sides are considered as functors from C×Set to Set, this isomorphism is a natural isomorphism.

The intuitive understanding of Yoneda's Lemma is that by studying the morphisms from an object to other objects, we can fully understand its properties. The profundity of Yoneda's Lemma lies in its revelation that objects in a category are not independently existing, but rather interconnected through their morphisms with other objects. This connection can be expressed in the language of natural transformations, providing a powerful tool for understanding the structure of categories and the relationships between objects. Yoneda's Lemma demonstrates that we can understand the nature of an object by examining the morphisms between it and other objects. In causal analysis, this can be interpreted as our ability to understand the essence of causal relationships by observing the morphisms from a causal relationship to a dataset representing that relationship, as well as the natural transformations from this causal relationship to another.

3.4.2 Data fusion theory proof based on Yoneda lemma

Definition 17: Let C be a locally small category, meaning that both the collections Ob(-) and Hom(-) defined on C are sets rather than proper classes. For any A in C and y in homc, set, if the transformation $Nat(Homc(-,z), Homc(-,y))$ is a natural transformation from the functor IC to the functor Y, then this transformation is called a data fusion.

Lemma 1: For the category class Ob(C) corresponding to a certain causal problem. For any A in C, let its data fusion be the natural transformation $Nat(hA,F)$, where F is a functor from C to the category of sets. Then this transformation is an isomorphism.

Proof: According to Definition 15, it can be seen that in C as an abstract representation, the elements in Ob(C) correspond to specific representations of C. Therefore, Ob(C) is a set rather than a class. Moreover, for any A in C, A represents a causal relationship, which is a category formed by functions, and this category is also a set. Therefore, C is a locally small category.

Let G be a preserving functor from category D to C x Set. For any B in D, $HomD/c = G$, by

Yoneda's Lemma and Definition 17, it is known that this transformation is an isomorphism.